\documentclass[journal]{IEEEtran}
\ifCLASSINFOpdf
\else
\fi
\usepackage{times}
\usepackage{graphicx} % more modern
\usepackage{epstopdf}
\usepackage{subfigure}

\usepackage{algorithm}
\usepackage{algorithmic}
\usepackage{array}
\usepackage{amsmath}
\usepackage{amsthm}
\usepackage{bm}
\usepackage{amsfonts}
\usepackage{amssymb}
\usepackage{multirow}
\usepackage{multicol}
\usepackage{booktabs}
\usepackage{multirow}
\usepackage{caption2}
\theoremstyle{plain}

\usepackage{color}

\makeatletter

\makeatletter

\newcommand{\norm}[1]{\ensuremath{\|#1\|}}
\newcommand{\onenorm}[1]{\ensuremath{\|#1\|_1}}
\newcommand{\twonorm}[1]{\ensuremath{\|#1\|_2}}
\newcommand{\twoonenorm}[1]{\ensuremath{\|#1\|_{2,1}}}
\newcommand{\gnorm}[1]{\ensuremath{\|#1\|_{G_1}}}

\newcommand{\fnorm}[1]{\ensuremath{\|#1\|_F}}

\begin{document}
\title{A Survey on Multi-View Clustering}

\author{Guoqing Chao$^1$, Shiliang Sun$^2$,  Jinbo Bi$^{1*}$% <-this % stops a space
\thanks{$^1$ Guoqing Chao and Jinbo Bi are with the Department of Computer Science and Engineering, University of Connecticut, Storrs, CT, USA (e-mail: guoqing.chao, jinbo.bi@uconn.edu).}% <-this % stops a space
\thanks{$^2$ Shiliang Sun is with the Department of Computer Science and Technology, East China Normal University, 3663 North Zhongshan Road, Shanghai 200062, PR China (email: slsun@cs.ecnu.edu.cn).}}% <-this %stops a space
\maketitle

\begin{abstract}
With advances in information acquisition technologies, multi-view data become ubiquitous. Multi-view learning has thus become more and more popular in machine learning and data mining fields. Multi-view unsupervised or semi-supervised learning, such as co-training, co-regularization has gained considerable attention. Although recently, multi-view clustering (MVC) methods have been developed rapidly, there has not been a survey to summarize and analyze the current progress. Therefore, this paper reviews the common strategies for combining multiple views of data and based on this summary we propose a novel taxonomy of the MVC approaches. We further discuss the relationships between MVC and multi-view representation, ensemble clustering, multi-task clustering, multi-view supervised and semi-supervised learning. Several representative real-world applications are elaborated. To promote future development of MVC, we envision several open problems that may require further investigation and thorough examination.
\end{abstract}
% \fi
\IEEEpeerreviewmaketitle
\section{Introduction}
\label{sec:introduction}

Clustering~\cite{Berkhin2002} is a paradigm to classify a sample of subjects into subgroups based on similarities among subjects. Clustering is  a fundamental task in machine learning, pattern recognition and data mining fields and it has widespread applications. Once subgroups can be obtained by clustering methods, many subsequent analytic tasks can be conducted to achieve different ultimate goals. Traditional clustering methods only use a single set of features or one information window of the subjects.  When multiple sets of features are available for each individual subject, how can these views are integrated to help identify essential grouping structure is a problem of our concern in this paper, which is often referred to as multi-view clustering. 

Multi-view data are very common in real-world applications in the big data era. For instance, a web page can be described by the words appearing on the web page itself and the words underlying all links pointing to the web page from other pages in nature.  In multimedia content understanding, multimedia segments can be simultaneously described by their video signals from visual camera and audio signals from voice recorder devices. The existence of such multi-view data raised the interest of multi-view learning~\cite{xu2013,sunshilang2013,zhaojing2017}, which has been extensively studied in the semi-supervised learning setting. For unsupervised learning, particularly, multi-view clustering, single view based clustering methods cannot make an effective use of the multi-view information in various problems. For instance, a multi-view clustering problem may require to identify clusters of subjects that differ in each of the data views. In this case, concatenating features from the different views into a single union followed by a single-view clustering method may not serve the purpose. It has no mechanism to guarantee that the resultant clusters differ from all of the views because a specific view of features may very likely be weighted much higher than other views in the feature union which renders the grouping is based only on one of the views.  Multi-view clustering has thus attracted more and more attentions in the past two decades, which makes it necessary and beneficial to summarize the state of the art and delineate open problems to guide future advancement.

We now give the definition of multi-view clustering (MVC). MVC is a machine learning paradigm to classify similar subjects into the same group and dissimilar subjects into different groups by combining the available multi-view feature information, and to search for consistent clusterings across different views. Similar to the categorization of clustering algorithms in~\cite{Berkhin2002}, we divide the existing MVC methods into two categories: generative (or model-based) approaches and discriminative (or similarity-based) approaches. Generative approaches try to learn the fundamental distribution of the data and use generative models to represent the data with each model representing one cluster. Discriminative approaches directly optimize an objective function that involves pairwise similarities to minimize the average similarity within clusters and to maximize the average similarity between clusters. Due to a large number of discriminative approaches, based on how they combine the multi-view information, we further divide them into five classes: (1) common eigenvector matrix (mainly multi-view spectral clustering), (2) common coefficient matrix (mainly multi-view subspace clustering), (3) common indicator matrix (mainly multi-view nonnegative matrix factorization clustering), (4) direct view combination (mainly multi-kernel clustering), (5) view combination after projection (mainly canonical correlation analysis (CCA)). The first three classes have a commonality that they share a similar structure to combine multiple views.
% So far, all the MVC methods we talked about only find one clustering solution. However, data can be often grouped and interpreted in many different ways. Therefore, we introduced several MVC methods that provide several solutions.

Research on MVC is motivated by the multi-view real applications, often the same ones that motivate to develop multi-view representation, multi-view supervised, and multi-view semi-supervised learning methods.  Therefore, the similarities and differences of these different learning paradigms are also worth discussing. An obvious commonality between them is that they all learn with multi-view information. However, their learning targets are different. Multi-view representation methods aim to learn a joint compact representation for subjects from all of the views whereas MVC aims to perform sample partitioning, and MVC is learned without any label information. In contrast, multi-view supervised and semi-supervised learning methods have access to all or part of the sample label information. Some of the view combination strategies in these related paradigms can be borrowed and adapted by MVC. In addition, the relationships among MVC, and ensemble clustering, and multi-task clustering are also elaborated in this review.

MVC has been applied to many scientific domains such as computer vision, natural language processing, social multimedia, bioinformatics, and health informatics. As far as what this paper is concerned, the methodology papers of MVC are published largely in top machine learning, pattern recognition, or data mining venues like the International Conference on Machine Learning (ICML)~\cite{vikas2008icml,Kumar2011ICML,hua2013icml,thorsten2001icml,dengyongicml2007,kamalika2009icml,nikhil2014icml,virginia2005icmlworkshop,javon2015,weiran2015icml,jiquan2011icml,donglin2010icml,junyuan2016icml,marc2017icml}, Neural Information Processing Systems (NIPS)~\cite{Kumar2011NIPS,tilman2005nips}, IEEE International Conference on Computer Vision and Pattern Recognition (CVPR)~\cite{jinlin2016cvpr,ron2007cvpr,nishant2016icpr,matthew2008cvpr}, International Conference on Computer Vision (ICCV)~\cite{abdelaziz2013iccv}, Association for the Advancement of Artificial Intelligence (AAAI)~\cite{handong2017aaai,qian2015aaai,xinwang2016aaai,rongkai2014aaai,cheng2015aaai}, International Joint Conference on Artificial Intelligence (IJCAI)~\cite{xiao2013ijcai,zhiqiang2017ijcai,feiping2017ijcai,chang2015ijcai,shaoyuan2014ijcai,handong2016ijcai,xianchao2015ijcai}, SIAM International Conference on Data Mining (SDM)~\cite{xiang2013sdm,bo2008sdm}, IEEE International Conference on Data Mining (ICDM)~\cite{wei2009icdm,grigorios2012icdm,guillaume2009icdm,weixiang2013icdm,ying2007icdm}. The journals that MVC methods are often present include IEEE Transactions on Pattern Analysis and Machine Intelligence (PAMI)~\cite{shi2012pami}, IEEE Transactions on Knowledge and Data Engineering (TKDE)~\cite{xiaojun2011tkde,chang2016tkde,xinhai2012tkde,hongfu2017tkde,xiaochao2016tkde}, IEEE Transactions on Cybernetics (TCYB)~\cite{yizhang2015tc,deng2014tc}, IEEE Transactions on Image Processing (TIP)~\cite{yang2015tip}, and IEEE Transactions on Neural Networks and Learning Systems (TNNLS)~\cite{grigorios2009tnn}. Although MVC has permeated into many fields and made great success in practice, there are still some open problems that limit its further advancement. We point out several open problems and hope they can be helpful to promote the development of MVC. With this survey, we hope that readers can have a more comprehensive version of the MVC development and what is beyond the current progress.  

The remainder of this paper is organized as follows. In section~\ref{sec:genera}, we review the existing generative methods for MVC. Section~\ref{sec:discri} introduces several classes of discriminative MVC methods. In Section~\ref{sec:relationships to related topics}, we analyze the relationships between MVC and several related topics. Section~\ref{sec:applications} presents the applications of MVC in different areas. In Section~\ref{sec:open problems}, we list several open problems in MVC research, which we aim to help advance the development of MVC. Finally, we make the conclusions.

\section{Generative Approaches}
\label{sec:genera}
Generative approaches aim to learn the generative models each of which is used to generate the data from a cluster. In most cases, generative clustering approaches are based on mixture models or constructed via expectation maximization (EM)~\cite{arthur1977jrss}. Therefore, we first introduce mixture models and EM algorithm. We will also review another popular single-view clustering model named convex mixture models (CMMs)~\cite{danial2008nips} which was extended to the multi-view case.
\subsubsection{Mixture Models and CMMs}
A generative approach assumes that data are sampled independently from a mixture model of multiple probability distributions. The mixture distribution can be written as
\begin{equation}\label{mixturedis}
p(\bm x|\bm \theta)=\sum_{k=1}^K\pi_k p(\bm x|\bm \theta_k),
\end{equation}
where $\pi_k$ is the prior probability of the $k$th component and satisfies $\pi_k\geq0,$ and $\sum_{k=1}^K\pi_k=1$, $\bm \theta_k$ is the parameter of the $k$th probability density model and $\bm \theta=\{(\pi_k,\bm \theta_k), k=1,2,\cdots,K\}$ is the parameter set of the mixture model. For instance, $\bm \theta_k=\{\bm \mu_k,\bm \Sigma_k\}$ for Gaussian mixture model.

The EM is a widely used algorithm for parameter estimation of the mixture models. Suppose that the observed data and unobserved data are denoted by $\bm X$ and $\bm Z$, respectively. $\{\bm X, \bm Z\}$ and $\bm X$ are called {\it complete data} and {\it incomplete data}. In the E (expectation) step, the posterior distribution $p(\bm Z|X,{\bm \theta}^{old})$ of the unobserved data is evaluated with the current parameter values ${\bm \theta}^{old}$. The E step calculates the expectation of the complete-data log likelihood evaluated for some general paramter value $\bm \theta$. The expectation, denoted by $Q(\bm \theta, {\bm \theta}^{old})$, is given by 
\begin{equation}\label{qfunction}
Q(\bm \theta, {\bm \theta}^{old})=\sum_{\bm Z}p(\bm Z|\bm X,{\bm \theta}^{old})\ \mbox{ln}\ p(\bm X,\bm Z|\bm \theta).
\end{equation}
The first item is the posterior distribution of the latent variables $\bm Z$ and the second one is the complete-data log likelihood. According to maximum likelihood estimation, the M step updates the parameters by maximizing the function~\eqref{qfunction}
\begin{equation}
\bm \theta=\mbox{arg}\ \underset{\bm \theta}{\mbox{max}}Q(\bm \theta, {\bm \theta}^{old}).
\end{equation}

Note that for clustering, $\bm X$ can be considered as the observed data while $\bm Z$ is the latent variable whose entry $z_{nk}$ indicates the $n$th data point comes from the $k$th component. Also note that the posterior distribution form used to be evaluated in E step and the expectation of the complete data log likelihood used to evaluate the parameters are different for different distribution assumptions. It can adopt Gaussian distribution and any other probability distribution form, which depends on the specific applications.

CMMs~\cite{danial2008nips} are simplified mixture models that can probabilistically assign data points to clusters after extracting the representative exemplars from the data set. By maximizing the log-likelihood, all instances compete to become the ``center" (representative exemplar) of the clusters. The instances corresponding to the components that received the highest priors are selected exemplars and then the remaining instances are assigned to the ``closest" exemplar. The priors of the components are the only adjustable parameters of a CMM.

Given a data set $\bm X={\bm x_1,\bm x_2,\cdots,\bm x_N}\in {\mathbb{R}}^{d\times N}$, the CMM distribution is $Q(\bm x)=\sum_{j=1}^N q_jf_j(\bm x)$, $\bm x\in {\mathbb{R}}^{d}$, where $q_j\geq0$ denotes the prior probability of the $j$th component that satisfies the constraint $\sum_{j=1}^N q_j=1$, and $f_j(\bm x)$ is an exponential family distribution, with its expected parameters equal to the $j$th data point. Due to the bijection relationship between the exponential families and Bregman divergences~\cite{arindam2005jmlr}, the exponential family $f_j(\bm x)=C_{\phi}(\bm x)\mbox{exp}(-\beta d_{\phi}(\bm x,\bm x_j))$ where $d_{\phi}$ denotes the Bregman divergence that calculates the component distribution, $C_{\phi}(\bm x)$ is independent of $\bm x_j$, and $\beta$ is a constant controlling the sharpness of the components.

The log-likelihood that needs to be maximized is given as $L(\bm X;\{q_j\}_{j=1}^N)=\frac{1}{N}\sum_{i=1}^N\mbox{log}\big(\sum_{j=1}^Nq_j f_j(\bm x_i)\big)=\frac{1}{N}\sum_{i=1}^N\mbox{log}\big(\sum_{j=1}^Nq_j e^{-\beta d_{\phi}(\bm x_i,\bm x_j)}\big)$+ const. If the empirical samples are equally drawn, i.e., the prior of drawing each example is $\hat{P}=1/N$, the log-likelihood can be equivalently expressed in terms of Kullback Leibler (KL) divergence between $\hat{P}$ and $Q(\bm x)$ as
\begin{equation}\label{cmmkl}
\begin{aligned}
&D(\hat{P}|Q)=-\sum_{i=1}^N\hat{P}(\bm x_i)\mbox{log} Q(\bm x_i)-\mathbb{H}(\hat{P})\\
&\quad\quad\quad\ =-L(\bm X;\{q_j\}_{j=1}^N)+\mbox{const}\\
\end{aligned}
\end{equation}
where $\mathbb{H}(\hat{P})$ is the entropy of the empirical distribution $\hat{P}(\bm x)$ which does not depend on the parameter $q_j$. Now, the problem is changed into minimizing~\eqref{cmmkl}, which is convex and can be solved by an iterative algorithm. In such an algorithm, the updating rule for prior probabilities is given by
\begin{equation}
q_j^{(t+1)}=q_j^{(t)}\sum_{i=1}^N\frac{\hat{P}(\bm x_i)f_j(\bm x_i)}{\sum_{j'=1}^N q_{j'}^{(t)}f_{j'}(\bm x_i)}.
\end{equation}

The data points are grouped into $K$ disjoint clusters by requiring the instances with the $K$ highest $q_j$ values to serve as exemplars and then assigning each of the remaining instances to an exemplar with which the instance has the highest posterior probability. Note that the clustering performance is affected by the value of $\beta$. In~\cite{danial2008nips} a reference value $\beta_0$ is determined using an empirical rule $\beta_0=N^2\mbox{log}N/\sum_{i,j=1}^N d_{\phi}(\bm x_i,\bm x_j)$ to identify a reasonable range of $\beta$,  which is around $\beta_0$. More details refer to Paper~\cite{danial2008nips}. 

\subsubsection{Multi-View Clustering Based on Mixture Models or EM Algorithm}\label{mvcmm}
The method in~\cite{bickel2004} assumes that the two views are independent, multinomial distribution is adopted for document clustering problem. It uses the two-view case as an example, and executes the M and E steps on each view and then interchange the posteriors  in two separate views in each iteration. The optimization process is terminated if the log-likelihood of observing the data does not reach a new maximum for a fixed number of iterations in each view. Two multi-view EM algorithm versions for finite mixture models are proposed in the paper~\cite{xing2005icapr}: the first version can be regarded as that it runs EM in each view and  then combines by adding the weighted probabilistic clustering labels generated in each view before each new EM iteration while the second version can be viewed as some probabilistic information fusion for components of the two views.

Specifically, based on the CMMs for single-view clustering, the multi-view version proposed in~\cite{grigorios2009icann} became much attractive because it can locate the global optimum and thus avoid the initialization and local optima problems of standard mixture models, which require multiple executions of the EM algorithms.

For multi-view CMMs, each $\bm x_i$ with $m$ views is denoted by $\{\bm x_i^1,\bm x_i^2,\cdots,\bm x_i^m\}$, $\bm x_i^v\in {\mathbb{R}}^{d^v}$, the mixture distribution for each view is given as  $Q^v(\bm x^v)=\sum_{j=1}^N q_jf_j^v(\bm x^v)=C_{\phi}(\bm x^v)\sum_{j=1}^N q_j e^{-{\beta}^v d_{\phi_v}(\bm x^v,\bm x_j^v)}$. To pursue a common clustering across all views, all $Q^v(\bm x^v)$ share the same priors. In addition, an empirical data set distribution $\hat{P}^v(\bm x^v)=1/N$, $\bm x^v\in\{\bm x_1^v,\bm x_2^v,\cdots,\bm x_N^v\}$, is associated with each view and the multi-view algorithm minimizes the sum of KL divergences between $\hat{P}^v(\bm x^v)$ and $Q^v(\bm x^v)$ across all views with the constraint $\sum_{j=1}^Nq_j=1$
\begin{equation}\label{mvcmms}
\begin{aligned}
&\underset{q_1,\cdots,q_N}{\mbox{min}}\sum_{v=1}^m D(\hat{P}^v|Q^v)\\
&\quad\quad\quad=\underset{q_1,\cdots,q_N}{\mbox{min}}-\sum_{v=1}^m\sum_{i=1}^N\hat{P}^v(\bm x_i^v)\mbox{log}Q^v(\bm x_i^v)-\sum_{v=1}^m\mathbb{H}(\hat{P}^v).\\
\end{aligned}
\end{equation}
which is straightforward to see that the optimized objective is convex, hence the global minimum can be found. The prior undate rule is given as follows:
\begin{equation}
q_j^{(t+1)}=\frac{q_j^{(t)}}{M}\sum_{v=1}^m\sum_{i=1}^N\frac{\hat{P}^vf_j^v(\bm x_i^v)}{\sum_{j'=1}^Nq_{j'}^{(t)}f_{j'}^v(\bm x_i^v)}.
\end{equation}
The prior $q_j$ associated with the $j$th instance is a measure of how likely this instance is to be an exemplar, taking all views into account.  The appropriate ${\beta}^v$ values are identified in the range of an empirically defined $\beta_0^v$ by $\beta_0^v=N^2\mbox{log}N/\sum_{i,j=1}^N d_{\phi_v}(\bm x_i^v,\bm x_j^v)$.
From Eq.~\eqref{mvcmms}, it can be found that all views contribute equally to the sum, without considering their different importance. To overcome this limitation, a weighted version of multi-view CMMs was proposed in~\cite{grigorios2010tnnls}.
\section{Discriminative Approaches}
\label{sec:discri}
Compared with generative approaches, discriminative approaches directly optimize the objective to seek for the best clustering solution rather than first modelling the samples then solving these models to determine clustering result. Directly focusing on the objective of clustering makes discriminative approaches gain more attentions and develop more comprehensively. Up to now, most of the existing MVC methods are discriminative approaches.
Based on how to combine multiple views, we categorize MVC methods into five main classes and introduce the representative works in each group.

The settings of MVC are introduced in~\ref{mvcmm} . The aim of MVC is to cluster the $N$ subjects into $K$ classes. That is, finally we will get a membership matrix $\bm H\in \mathbb{R}^{N\times K}$ to indicate which subjects are in the same group while others in other classes, the sum of each row entries of $\bm H$ should be 1 to make sure each row is a probability. If only one entry of each row is 1 and all others are 0, it is the so-called hard clustering otherwise it is soft clustering.
\subsection{Common Eigenvector Matrix (Mainly Multi-View Spectral Clustering)}
\label{subsec:multi-view spectral clustering}
This group of MVC methods are based on a commonly used clustering technique spectral clustering. Since spectral clustering hinges crucially on the construction of the graph Laplacian and the resulting eigenvectors reflect the grouping structure of the data, this group of MVC methods guarantee to get a common clustering results by assuming that all the views share the same or similar eigenvector matrix. There are two representative methods: co-training spectral clustering~\cite{Kumar2011ICML} and co-regularized spectral clustering~\cite{Kumar2011NIPS}. Before discussing them, we will introduce spectral clustering~\cite{Andrew2001NIPS} first.
\subsubsection{Spectral Clustering}
Spectral clustering is a clustering technique that utilizes the properties of graph Laplacian where the graph edges denote the similarities between data points and solve a relaxation of the normalized min-cut problem on the graph~\cite{jianbo2000PAMI}. Compared with other widely used methods such as the k-means method that only fits the spherical shaped clusters, spectral clustering can apply to arbitrary shaped clusters and demonstrate good performance.

Given $\bm G=(\bm V,\bm E)$ as a weighted undirected graph with vertex set $\bm V={v_1,\cdots,v_N}$. The data adjacency matrix of the graph is defined as $\bm W$ whose entry $w_{ij}$ represents the similarity of two vertices $v_i$ and $v_j$. If $w_{ij}=0$ it means that the vertices $v_i$ and $v_j$ are not connected. Apparently $\bm W$ is symmetric because $\bm G$ is an undirected graph. The degree matrix $\bm D$ is defined as the diagonal matrix with the degrees $d_1,\cdots,d_N$ of each vertex on the diagonal, where $d_i=\sum_{j=1}^N w_{ij}$. Generally, the graph Laplacian is $\bm D-\bm W$ and the normalized graph Laplacian is $\tilde{\bm L}=\bm D^{-1/2}\bm (\bm D-\bm W) \bm D^{-1/2}$. In many spectral clustering works e.g.~\cite{Andrew2001NIPS,Kumar2011ICML,Kumar2011NIPS}, $\bm L=\bm D^{-1/2}\bm W\bm D^{-1/2}$ is also used to change a minimization problem~\eqref{scmin} into a maximization problem~\eqref{sc} since $\bm L=\bm I-\tilde{\bm L}$ where $\bm I$ is the identity matrix. Following the same terminology adopted in~\cite{Andrew2001NIPS,Kumar2011ICML,Kumar2011NIPS}, we will name both $\bm L$ and $\tilde{\bm L}$ as normalized graph Laplacians afterwards. Now the single-view spectral clustering approach can be formulated as follows:
\begin{equation}
\begin{split}\label{sc}
\left\{\begin{matrix}
&\underset{\bm U\in \mathbb{R}^{N\times K}}{\mbox{max}} tr({\bm U}^{\mathrm{T}}\bm L\bm U)\\
&s.t.\quad {\bm U}^{\mathrm{T}}\bm U=\bm I,
\end{matrix}\right.
\end{split}
\end{equation}
which is also equivalent to the following problem:
\begin{equation}
\begin{split}\label{scmin}
\left\{\begin{matrix}
&\underset{\bm U\in \mathbb{R}^{N\times K}}{\mbox{min}} tr({\bm U}^{\mathrm{T}}\tilde{\bm L}\bm U)\\
&s.t.\quad {\bm U}^{\mathrm{T}}\bm U=\bm I,
\end{matrix}\right.
\end{split}
\end{equation}
where $tr$ denotes the trace norm of a matrix. The rows of matrix $\bm U$ are the embeddings of the data points, which can be fed into the k-means to obtain the final clustering results. A version of the Rayleigh-Ritz theorem in~\cite{helmut1997} shows that the solution of the above optimization problem is given by choosing $\bm U$ as the matrix containing, respectively, the largest or smallest $K$ eigenvectors of $\bm L$ or $\tilde{\bm L}$ as columns.
To understand the spectral clustering method better, we outline a commonly used algorithm~\cite{Andrew2001NIPS} as follows:
\begin{itemize}
  \item  Construct the adjacency matrix $\bm W$.
  \item  Compute the normalized Laplacian matrix $\bm L=\bm D^{-1/2}\bm W\bm D^{-1/2}$.
  \item  Calculate the eigenvectors of $\bm L$ and stack the top $K$ eigenvectors as the columns to construct a $N\times K$ matrix $\bm U$.
  \item  Normalize each row of $\bm U$ to obtain $\bm U_{sym}$.
  \item  Run the k-means algorithm to cluster the row vectors of $\bm U_{sym}$.
  \item  Assign subject $i$ to cluster $k$ if the $i$th row of $\bm U_{sym}$ is assigned to cluster $k$ by the k-means algorithm.
\end{itemize}

Apart from the symmetric normalization operator $\bm U_{sym}$, another normalization operator $\bm U_{lr}=\bm D^{-1}\bm W$ is also commonly used. Refer to~\cite{ulrike2007} for more details about spectral clustering.

\subsubsection{Co-Training Multi-View Spectral Clustering}
For semi-supervised learning, co-training with two views has been a widely recognized idea when both labeled and unlabeled data are available. It assumes that the predictive models constructed in each of the two views will lead to the same labels for the same sample with high probability. There are two main assumptions to guarantee the success of co-training: (1) \textit{Sufficiency}: each view is sufficient for sample classification on its own, (2) \textit{Conditional independence}: the views are conditionally independent given the class labels. In the original co-training algorithm~\cite{blum1998combining},  two initial predictive functions $f_1$ and $f_2$ are trained in each view using the labeled data, then the following steps are repeatedly performed: the most confident examples predicted by $f_1$ are added to the labeled set to train $f_2$ and vice versa, then $f_1$ and $f_2$ are re-trained on the enlarged labeled datasets. It can be shown that after a number of iterations, $f_1$ and $f_2$ will agree with each other on labels.

For co-training multi-view spectral clustering, the motivation is similar: the clustering result in all views should agree. In spectral clustering, the eigenvectors of the graph Laplacian encode the discriminative information of the clustering. Therefore, co-training multi-view spectral clustering~\cite{Kumar2011ICML} uses the eigenvectors of the graph Laplacian in one view to cluster samples and then use the clustering result to modify the graph Laplacian in the other view.

Each column of the similarity matrix (also called the adjacency matrix) $\bm W_{N\times N}$ can be considered as a $N$-dimensional vector that indicates the similarities of $i$th point with all the points in the graph. Since the largest $K$ eigenvectors have the discriminative information for clustering, the similarity vectors can be projected along those directions to retain the discriminative information for clustering and throw away the within cluster details that might confuse the clustering. After that, the projected information is projected back to the original $N$-dimensional space to get the modified graph. Due to the orthogonality of the projection matrix, the inverse projection is equivalent to the transpose operation.

To make the co-training spectral clustering algorithm clear, we borrowed Algorithm 1 from~\cite{Kumar2011ICML}. Note that a symmetrization operator $sym$ on a matrix $\bm S$ is defined as $sym\ (\bm S)=(\bm S+\bm S^\mathrm{T})/2$ in Algorithm 1.
\begin{algorithm}[h]
   \caption{$~~$Co-training Multi-View Spectral Clustering}
   \label{alg:co-traing sc}
\begin{algorithmic}
    \STATE {\bfseries Input:} Similarity matrices for two views: $\bm W^{(1)}$ and $\bm W^{(2)}$.
	\STATE {\bfseries Output:} Assignments to $K$ clusters.
	\STATE {\bfseries Initialize:} $\bm L^{(v)}={\bm D^{(v)}}^{(-1/2)}\bm L^{(v)}{\bm D^{(v)}}^{(-1/2)}$ for $v=1, 2$,
${\bm U^{(v)}}^0=\underset{\bm U\in \mathbb{R}^{N\times K}}{\mbox{argmax}} tr({\bm U}^{\mathrm{T}}\bm L^{(v)}\bm U)\ \
s.t.\quad {\bm U}^{\mathrm{T}}\bm U=I$ for $v=1, 2$.
    \STATE {\bfseries for} i=1 to T do
    \STATE 1. $\bm S^{(1)}=sym\ ({\bm U^{(2)}}^{i-1}{{\bm U^{(2)}}^{i-1}}^{\mathrm{T}}\bm W^{(1)})$
	\STATE 2. $\bm S^{(2)}=sym\ ({\bm U^{(1)}}^{i-1}{{\bm U^{(1)}}^{i-1}}^{\mathrm{T}}\bm W^{(2)})$
	\STATE 3. Use $\bm S^{(1)}$ and $\bm S^{(2)}$ as the new graph similarities and compute the graph Laplacians. Solve for the largest $K$ eigenvectors to obtain ${\bm U^{(1)}}^{i}$ and ${\bm U^{(2)}}^{i}$
	\STATE {\bfseries end for}
    \STATE  4: Normalize each row of ${\bm U^{(1)}}^{i}$ and ${\bm U^{(2)}}^{i}$.
    \STATE  5: Form matrix $\bm V={\bm U^{(v)}}^{i}$, where $v$ is the most informative view a priori. If there is no prior knowledge on the view informativeness, matrix $\bm V$ can also be set to be column-wise concatenation of the two ${\bm U^{(v)}}^{i}$s.
    \STATE  6: Assign example $j$ to cluster $K$ if the $j$th row of $\bm V$ is assigned to cluster $K$ by the k-means algorithm.

\end{algorithmic}
\end{algorithm}

\subsubsection{Co-Regularized Multi-View Spectral Clustering}
Co-regularization is an effective technique in semi-supervised multi-view learning. The core idea of co-regularization is to minimize the distinction between the predictor functions of two views acts as one part of the objective function. However, there are no predictor functions in unsupervised learning like clustering, so how to implement the co-regularization idea in clustering problem? Co-regularized multi-view spectral clustering~\cite{Kumar2011NIPS} adopted the eigenvectors of graph Laplacian to play the similar role of predictor functions in semi-supervised learning scenario and proposed two co-regularized clustering approaches.

Let $\bm U^{(s)}$ and $\bm U^{(t)}$ be the eigenvector matrices corresponding to any pair of view graph Laplacians $\bm L^{(s)}$ and $\bm L^{(t)}$ ($1 \leq s, t \leq m, s\neq t$). The first version uses a pair-wise co-regularization criteria that enforces $\bm U^{(s)}$ and $\bm U^{(t)}$ as close as possible. The measure of clustering disagreement between the two views $s$ and $t$ is $D(\bm U^{(s)},\bm U^{(t)})=\fnorm{\frac{\bm K^{(s)}}{\fnorm{K^{(s)}}^2}-\frac{\bm K^{(t)}}{\fnorm{\bm K^{(t)}}^2}}^2$, where $\bm K^{(s)}=\bm U^{(s)}{\bm U^{(s)}}^{\mathrm{T}}$ using linear kernel is the similarity matrix of $\bm U^{(s)}$. Since $\fnorm{\bm K^{(s)}}^2=K$, where $K$ is the number of the clusters, disagreement between the clustering solutions in the two views can be measured by $D(\bm U^{(s)},\bm U^{(t)})=-tr(\bm U^{(s)}{\bm U^{(s)}}^{\mathrm{T}}\bm U^{(t)}{\bm U^{(t)}}^{\mathrm{T}})$. Integrating the measure of the disagreement between any pair of views into the spectral clustering objective function, the pair-wise co-regularized multi-view spectral clustering can be formed as the following optimization problem:
\begin{equation}
\begin{split}\label{pwsc}
\left\{\begin{matrix}
&\underset{\bm U^{(1)},\bm U^{(2)},\cdots,\bm U^{(m)}\in \mathbb{R}^{N\times K}}{\mbox{max}} \sum_{s=1}^{m}({\bm U^{(s)}}^{\mathrm{T}}\bm L^{(s)}\bm U^{(s)})\\
&+\sum_{1 \leq s, t \leq m, s\neq t}\lambda \ tr(\bm U^{(s)}{\bm U^{(s)}}^{\mathrm{T}}\bm U^{(t)}{\bm U^{(t)}}^{\mathrm{T}})
\\
&s.t.\quad {\bm U^{(s)}}^{\mathrm{T}}\bm U^{(s)}=\bm I, \ \forall 1\leq s\leq m.
\end{matrix}\right.
\end{split}
\end{equation}
The hyperparameter $\lambda$ is used to trade-off the spectral clustering objectives and the spectral embedding disagreement terms. After the embeddings are obtained, each $\bm U^s$ can be fed for k-means clustering method, the final results are marginally different.

The second version named centroid-based co-regularization enforces the eigenvector matrix from each view to be similar by regularizing them towards a common consensus eigenvector matrix. The corresponding optimization problem is formulated as
\begin{equation}
\begin{split}\label{cbsc}
\left\{\begin{matrix}
&\underset{\bm U^{(1)},\bm U^{(2)},\cdots,\bm U^{(m)}, \bm U^{*}\in \mathbb{R}^{N\times K}}{\mbox{max}} \sum_{s=1}^{m}({\bm U^{(s)}}^{\mathrm{T}}\bm L^{(s)}\bm U^{(s)})\\
&+\lambda_s\sum_{s=1}^m tr(\bm U^{(s)}{\bm U^{(s)}}^{\mathrm{T}}\bm U^{(*)}{\bm U^{(*)}}^{\mathrm{T}})
\\
&s.t.\quad {\bm U^{(s)}}^{\mathrm{T}}\bm U^{(s)}=\bm I, \ \forall 1\leq s\leq m,\quad {\bm U^{*}}^{\mathrm{T}}\bm U^{*}=\bm I.
\end{matrix}\right.
\end{split}
\end{equation}

Compared with pairwise co-regularized version, centroid-based multi-view clustering does not need to combine the obtained eigenvector matrices of all views to run k-means. However, the centroid-based version possesses one potential drawback: the noisy views could potentially affect the optimal eigenvectors as it depends on all the views.

Cai et. al.~\cite{xiao2011} used a common indicator matrix across the views to perform multi-view spectral clustering and derived a formulation similar to the centroid-based co-regularization method. The main difference is that ~\cite{xiao2011} used $tr(({\bm U}^{(*)}-{\bm U}^{(s)})^{\mathrm{T}}({\bm U}^{(*)}-{\bm U}^{(s)}))$ as the disagreement measure between each view eigenvector matrix and the common eigenvector matrix while co-regularized multi-view spectral clustering ~\cite{Kumar2011NIPS} adopted $tr(\bm U^{(s)}{\bm U^{(s)}}^{\mathrm{T}}\bm U^{(*)}{\bm U^{(*)}}^{\mathrm{T}})$. The optimization problem~\cite{xiao2011} is formulated as
\begin{equation}
\begin{split}\label{osc}
\left\{\begin{matrix}
&\underset{ \bm U^{(s)},s=1,2\cdots,m, \bm U^{*} }{\mbox{max}} \sum_{s=1}^{m}({\bm U^{(s)}}^{\mathrm{T}}\bm L^{(s)}\bm U^{(s)})\\
&+\lambda\sum_1^m tr(({\bm U}^{*}-{\bm U}^{(s)})^{\mathrm{T}}({\bm U}^{*}-{\bm U}^{(s)}))\\
&s.t.\quad \bm U^{*}\geq 0, \quad {\bm U^{*}}^{\mathrm{T}}\bm U^{*}=I.\\
\end{matrix}\right.
\end{split}
\end{equation}
where $\bm U^{*}\geq 0$ makes $\bm U^{*}$ become the final cluster indicator matrix. Different from general spectral clustering that get eigenvector matrix first and then run clustering (such as k means that is sensitive to initialization condition) to assign clusters, Cai et al.~\cite{xiao2011} directly solves the final cluster indicator matrix, thus it will be more robust to the initial condition.

\subsubsection{Others}
Besides the two representative multi-view spectral clustering methods discussed above, Wang et al.~\cite{xiang2013sdm} enforces a common eigenvector matrix across the views and formulates a multi-objective problem which is then solved using Pareto optimization.

A relaxed kernel k means can be shown to be equivalent to spectral clustering, see the following Subsection~\ref{kkmeansc}, Ye et al.~\cite{yongkai2016icpr} proposes a co-regularized kernel k-means for multi-view clustering. With a multi-layer Grassmann manifold interpretation, Dong et al.~\cite{xiaowen2014tsp} obtains  the same formulation with the pair-wise co-regularized multi-view spectral clustering. 
\subsection{Common Coefficient Matrix (Mainly Multi-View Subspace Clustering)}
\label{subsec:multi-view subspace clustering}
In many practical applications, even though the given data are high dimensional, the intrinsic dimension of the problem is often low. For example, the number of pixels in a given image can be large, yet only a few parameters are used to describe the appearance, geometry and dynamics of a scene. This motivates the development of finding the underlying low dimensional subspace. In practice, the data could be sampled from multiple subspaces. Subspace clustering~\cite{rene2011} is the technique to find the underlying subspaces and then cluster the data points correctly according to the identified subspaces.
\subsubsection{Subspace clustering}
Subspace clustering uses the self-expressiveness property~\cite{ehsan2013} of the data samples, i.e., each sample can be represented by a linear combination of few other data samples. The classic subspace clustering formulation is given as follows:
\begin{equation}
\begin{split}
\bm X=\bm X\bm Z+\bm E
\end{split}
\end{equation}
where $\bm Z=\{z_1,z_2,\cdots,z_N\}\in {\mathbb {R}}^{N\times N}$ is the subspace coefficient matrix (representation matrix), and each $z_i$ is the representation of the original data point $x_i$ based on the subspace. $\bm E\in {\mathbb {R}}^{N\times N}$ is the noise matrix.

The subspace clustering can be formulated as the following optimization problem:
\begin{equation}
\begin{split}
\left\{\begin{matrix}
&\underset{\bm Z}{\mbox{min}} \fnorm{\bm X-\bm X\bm Z}^2\\
&s.t.\quad \bm Z(i,i)=0, {\bm Z}^{\mathrm{T}}\bm 1=\bm 1.
\end{matrix}\right.
\end{split}
\end{equation}
The constraint $\bm Z(i,i)=0$ is to avoid the case that a data point is represented by itself while ${\bm Z}^{\mathrm{T}}\bm 1=\bm 1$ denotes that the data point lies in a union of affine subspaces. The nonzero elements of $z_i$ correspond to data points from the same subspace.

After getting the subspace representation $\bm Z$, the similarity matrix $\bm W=\frac{|\bm Z|+|{\bm Z}^{\mathrm{T}}|}{2}$ can be obtained to further construct the graph Laplacian and then run spectral clustering on that graph Laplacian to get the final clustering results.
\subsubsection{Multi-View Subspace Clustering}
With multi-view information, each subspace representation $\bm Z_v$ can be obtained from each view. To get a consistent clustering result from multiple views, Yin et al.~\cite{qiyue2015} shares the common coefficient matrix by enforcing the coefficient matrices from each pair of views as similar as possible. The optimization problem is formulated as
\begin{equation}
\begin{split}\label{mvssc}
\left\{\begin{matrix}
&\underset{{\bm Z}^{(s)},s=1,2,\cdots,m}{\mbox{min}} \sum_{s=1}^{m}\fnorm{\bm X^{(s)}-\bm X^{(s)}\bm Z^{(s)}}^2\\
&\quad\quad+\alpha\sum_{s=1}^{m}\onenorm{\bm Z^{(s)}}+\beta\sum_{1\leq s\le t}\onenorm{\bm Z^{(s)}-\bm Z^{(t)}}\\
&s.t.\quad \mbox{diag}(\bm Z^{(s)})=0,\quad \forall s\in\{1,2,\cdots,m\}.
\end{matrix}\right.
\end{split}
\end{equation}
where \onenorm{\bm Z^{(s)}-\bm Z^{(t)}} is the $l_1$-norm based pairwise co-regularization constraint that can alleviate the noise problem. $\onenorm{\bm Z}$ is used to enforce sparse solution. $\mbox{diag}(\bm Z)$ denotes the diagonal elements of matrix $\bm Z$, and the zero constraint is used to avoid trivial solution (each data point represents by itself).

Wang et al.~\cite{yang2015} enforced the similar idea to combine multi-view information. Apart from that, it adopted a multi-graph regularization with each graph Laplacian regularization characterizing the view-dependent non-linear local data similarity. At the same time, it assumes that the view-dependent representation is low rank and sparse and considers the sparse noise in the data. Wang et al.~\cite{yang2015tip} proposed an angular based similarity to measure the correlation consensus in multiple views and obtained a robust subspace clustering for multi-view data.  Different from the above approaches, These three works~\cite{shaoyuan2014ijcai,handong2016ijcai,qiyue2015cikm} adopted general nonnegative matrix factorization formulation but shared a common representation matrix for the samples with both views and kept each view representation matrix specific. Zhao et al.\cite{handong2017aaai} adopted a deep semi-nonnegative matrix factorization to perform multi-view clustering, in the last layer a common coefficient matrix is enforced to exploit the multi-view information.

\subsection{Common Indicator Matrix (Mainly Multi-View Nonnegative Matrix Factorization Clustering)}
\label{subsec:multi-view nonnegative matrix factorization clustering}
\subsubsection{Nonnegative Matrix Factorization (NMF)}
For a nonnegative data matrix $\bm X\in {\mathbb {R}}_{+}^{d\times N}$, Nonnegative Matrix Factorization~\cite{daniel1999} seeks two nonnegative matrix factors $\bm U\in {\mathbb {R}}_{+}^{d\times K}$ and $\bm V\in {\mathbb {R}}_{+}^{N\times K}$ such that their product is a good approximation of $\bm X$:
\begin{equation}
\begin{split}
\bm X\approx \bm U\bm V^{\mathrm{T}},
\end{split}
\end{equation}
where $K$ denotes the desired reduced dimension (for clustering, it is the number of clusters), $\bm U$ is the basis matrix, and $\bm V$ is the indicator matrix.

Due to the nonnegative constraints, a widely known property of NMF is that it can learn a part-based representation. It is intuitive and meaningful in many applications such as in the face recognition~\cite{daniel1999}. The samples in many of these applications e.g., information retrieval~\cite{daniel1999} and pattern recognition~\cite{wei2003}  can be explained as additive combinations of nonnegative basis vectors. The NMF has been applied successfully to cluster analysis and shown the state-of-the-art performance~\cite{daniel1999,jean2004}.

\subsubsection{Multi-View Clustering based on NMF}
To combine multi-view information in the NMF framework, Akata et al.~\cite{zeynep2011} enforces a common indicator matrix in the NMF among different views to perform multi-view clustering. However, the indicator matrix $\bm V^{(v)}$ might not be comparable at the same scale. In order to keep the clustering solutions across different views meaningful and comparable, Liu et al.~\cite{jialu2013} enforces a constraint to push each view-dependent indicator matrix towards a common indicator matrix, which leads to another normalization constraint inspired by the connection between NMF and probability latent semantic analysis. The final optimization problem is formulated as:
\begin{equation}
\begin{split}\label{mvnmf}
\left\{\begin{matrix}
&\underset{{\bm U}^{(v)},{\bm V}^{(v)},v=1,2,\cdots,m}{\mbox{min}} \sum_{v=1}^{m}\fnorm{\bm X^{(v)}-\bm U^{(v)}\bm V^{(v)}}^2\\
&\quad+\sum_{v=1}^{m}\lambda_v\fnorm{\bm V^{(v)}-\bm V^{*}}^2\\
&s.t.\quad \forall 1\leq k\leq K, \onenorm{\bm U^{(v)}_{.,k}}=1, \bm U^{(v)},\bm V^{(v)},\bm V^{(*)}\geq 0.
\end{matrix}\right.
\end{split}
\end{equation}
The constraint $\onenorm{\bm U^{(v)}_{.,k}}=1$ is used to guarantee $\bm V^{(v)}$ within the same range for different $v$ such that the comparison between the view-dependent indicator matrix $\bm V^{(v)}$ and the consensus indicator matrix $\bm V^{(*)}$ is reasonable.
After obtaining the consensus matrix $\bm V^*$, the cluster label of data point $i$ can be computed as $arg max_k \bm V_{i,k}^*$.

\subsubsection{Multi-View K-Means}
The k-means clustering method can be formulated using NMF by introducing an indicator matrix $\bm H$. The NMF formulation of k-means clustering is
\begin{equation}
\begin{split}\label{nmfkm}
\left\{\begin{matrix}
&\underset{\bm H,\bm G}{\mbox{min}}\fnorm{\bm X^{\mathrm{T}}-\bm H\bm G^{\mathrm{T}}}^2\\
& s.t.\quad \bm H_{i,k}\in\{0,1\}, \sum_{k=1}^{K}\bm H_{i,k}=1, \forall i=1,2,\cdots,N
\end{matrix}\right.
\end{split}
\end{equation}
where the columns of $\bm G \in \mathbb {R}^{d\times K}$ give the cluster centroids.

Because the k-means algorithm does not suffer expensive computation cost such as that required by eigen-decomposition, it can be a good choice for large scale data clustering. To deal with large scale multi-view data, Cai et al.~\cite{xiao2013ijcai} proposed a multi-view k-means clustering method by adopting a common indicator matrix across different views. The optimization problem is formulated as follows:
\begin{equation}
\begin{split}\label{mvkm}
\left\{\begin{matrix}
&\underset{{\bm G}^{(v)},{\alpha}^{(v)},\bm H}{\mbox{min}} \sum_{v=1}^{m}(\alpha^{(v)})^{\gamma}\twoonenorm{{\bm X^{(v)}}^{\mathrm{T}}-\bm H\bm G^{\mathrm{T}}}\\
& s.t.\quad \bm H_{i,k}\in\{0,1\}, \sum_{k=1}^{K}\bm H_{i,k}=1, \sum_{v=1}^{m}{\alpha}^{(v)}=1
\end{matrix}\right.
\end{split}
\end{equation}
where ${\alpha}^{(v)}$ is the weight for the $v$-th view and $\gamma$ is the parameter to control the weights distribution. By learning the weights $\alpha$ for different views, the important views will get large weight during multi-view clustering.

\subsubsection{Others}
As mentioned earlier, there are generally two steps in subspace clustering: find a subspace representation and then run spectral clustering on the graph Laplacian computed from the subspace representation. To identify consistent clusters from different views, Gao et al.~\cite{hongchang2015} merged these two steps in subspace clustering and enforced a common indicator matrix across different views. The formulation is given as follows:
\begin{equation}
\begin{split}\label{mvssp}
\left\{\begin{matrix}
&\underset{{\bm Z}^{(v)},{\bm E}^{(v)},\bm H}{\mbox{min}} \sum_{v=1}^{m}\fnorm{\bm X^{(v)}-\bm X^{(v)}\bm Z^{(v)}-\bm E^{(v)}}^2\\
&\quad\quad+\lambda_1 tr(\bm H^\mathrm{T}(\bm D^{(v)}-\bm W^{(v)})\bm H)+\lambda_2\sum_{v=1}^{m}\onenorm{\bm E^{(v)}}\\
&s.t.\quad {\bm Z^{(v)}}^{\mathrm{T}},\bm Z^{(v)}(i,i)=\bm I, {\bm H}^{\mathrm{T}}\bm H=\bm I\\
\end{matrix}\right.
\end{split}
\end{equation}
where $\bm Z^{(v)}$ is the subspace representation matrix of the $v$-th view, $\bm W^{(v)}=\frac{|\bm Z^{(v)}|+|{\bm Z^{(v)}}^{\mathrm{T}}|}{2}$, $\bm D^{(v)}$ is a diagonal matrix with diagonal elements defined as $d_{v_{i,i}}=\sum_{j}w_{v_{i,j}}$, and $\bm H$ is the common indicator matrix which indicates a unique cluster assignment for all the views. Although this multi-view subspace clustering method is based on subspace clustering, it does not enforce a common coefficient matrix $\bm Z$ but uses a common indicator matrix for different views. We thus categorize it into this group. {\bf I moved this one here because this is not NMF.}

Wang et al.~\cite{hua2013icml} integrates multi-view information via a common indicator matrix and simultaneously select features for different data clusters by formulating the problem as follows:
\begin{equation}
\begin{split}\label{mvcss}
\left\{\begin{matrix}
&\underset{{\bm H}^{\mathrm{T}}{\bm H}=\bm I,\bm W}{\mbox{min}} \fnorm{{\bm X}^{\mathrm{T}}\bm W+\bm 1_N {\bm b}^{\mathrm{T}}-\bm H}\\
& +\gamma_1\gnorm{\bm W}+\gamma_2\twoonenorm{\bm W}
\end{matrix}\right.
\end{split}
\end{equation}
where $\bm X=\{\bm x_1,\bm x_2,\cdots,\bm x_N\}\in\mathbb {R}^{d\times N}$, but here each $\bm x_i$ includes the features from all the $m$ views and each view has $d_j$ features such that $d=\sum_{j=1}^m d_j$. The coefficient matrix $\bm W=[w_1^1,\cdots,w_K^1;\cdots,\cdots,\cdots,;w_1^m,\cdots,w_K^m]\in {\mathbb {R}}^{d\times K}$ contains the weights of each feature for $K$ clusters, $\bm b\in \mathbb {R}^{K\times 1}$ is the intercept vector, $\bm 1_N$ is $N$-element constant vector of ones, and $\bm H=[h_1,\cdots,h_N]^{\mathrm{T}}\in \mathbb {R}^{N\times K}$ is the cluster (assignment) indicator matrix. The regularizer $\gnorm{\bm W}=\sum_{i=1}^K\sum_{j=1}^m\twonorm{w_i^j}$ is the group $l_1$ regularization to evaluate the importance of an entire view's features as a whole for a cluster whereas $\twoonenorm{\bm W}=\sum_{i=1}^d\twonorm{w^i}$ is the $l_{2,1}$ norm to select individual features from all views that are important for all clusters.

In ~\cite{derek2009ecmlpkdd}, a matrix factorization approach was adopted to reconcile the clusters arisen from the individual views. Specifically, a matrix that contains the partition indicator of every individual view is created and then decomposed into two matrices: one showing the contribution of individual groupings to the final multi-view clustering, called meta-clusters, and the other showing the assignment of instances to the meta-clusters.
Tang et al.~\cite{wei2009icdm} treated multi-view clustering as clustering with multiple graphs, each of which is approximated by matrix factorization with two factors: a graph-specific factor and a factor common to all graphs. Qian et al.~\cite{Qian2016DoubleCN} required each view's indicator matrix as close as possible to a common indicator matrix and employed the Laplacian regularization to maintain the latent geometric structure of the views simultaneously. 

Besides using a common indicator matrix, ~\cite{weixiang2015,yumeng2016pr,weixiang2016bigdata} introduced a weight matrix to indicate whether there are missing entries so that it can tackle the missing value problem. The multi-view self-paced clustering method~\cite{chang2015ijcai} takes the complexities of the samples and views into consideration to alleviate the local minima problem. Tao et al.~\cite{zhiqiang2017ijcai} enforces a common indicator matrix and seeks for the consensus clustering among all the views in an ensemble way.
Another method that utilizes a common indicator matrix to combine multiple views~\cite{jinlin2016cvpr} employed the linear discriminant analysis idea and automatically weighed different views. For graph-based clustering methods, the similarity matrix for each view is obtained first, Nie et al.~\cite{feiping2017ijcai} assumes a common indicator matrix and then solves the problem by minimizing the differences between the common indicator matrix and each similarity matrix.

\subsection{Direct Combination (Mainly Multi-Kernel Based Multi-View Clustering)}
\label{subsec:multi-kernel clustering}
Besides the methods that share some structure among different views, direct view combination via a kernel is another common way to perform multi-view clustering. A natural approach is to define a kernel for each view and then combine these kernels in a convex combination~\cite{thorsten2001icml,tong2006kdd,guoqing2016ida}.
\subsubsection{Kernel Functions and Kernel Combination Methods}Kernel is a trick to learn nonlinear problem just by linear learning algorithm, since kernel function $K:\mathcal{X}\times\mathcal{X}\rightarrow\mathbb {R}$ can directly give the inner products in feature space without explicitly defining the nonlinear transformation $\phi$.
There are some common kernel functions as follows:
\begin{itemize}
  \item Linear kernel: $\ K(\bm x_i,\bm x_j)= (\bm x_i\cdot\bm x_j)$,
  \item Polynomial kernel: $ K(\bm x_i,\bm x_j)= (\bm x_i\cdot\bm x_j+1)^d$,
  \item Gaussian kernel (Radial basis kernel): $ K(\bm x_i,\bm x_j)= (\mbox{exp}\big(-\frac{\norm{\bm x_i-\bm x_j}^2}{2{\sigma}^2})$,
  \item Sigmoid kernel: $ K(\bm x_i,\bm x_j)= (\mbox{tanh}(\eta\bm x_i\cdot\bm x_j+\nu))$.
\end{itemize}

Kernel functions in a reproducing kernel Hilbert space (RKHS) can be viewed as similarity functions~\cite{vert2004} in a vector space, so we can use a kernel as a non-Euclidean similarity measure in the spectral clustering and kernel k-means methods. There have been some works on multi-kernel learning for clustering~\cite{bin2009siam,hong2009icann,hamed2006nips}, however, they are all for single-view clustering. If a kernel is derived from each view, and different kernels are combined elaborately to deal with the clustering problem, it will become the multi-kernel learning method for multi-view clustering.  Obviously, multi-kernel learning~\cite{gert2004jmlr,francis2004icml,soren2005nips,mehmet2008icml} can be considered as the most important part in this kind of multi-view clustering methods. There are three main categories of methods for combining multiple kernels~\cite{mehmet2011jmlr}:
\begin{itemize}
\item Linear combination: It includes two basic subcategories: unweighted sum $K(\bm x_i,\bm x_j)= \sum_{v=1}^m k_v(\bm x_i^v, \bm x_j^v)$ and weighted sum $K(\bm x_i,\bm x_j)= \sum_{v=1}^m w_v^q k_v(\bm x_i^v, \bm x_j^v)$ where $w_v\in\mathbb{R_+}$ denotes the kernel weight for the $v$th view and $\sum_{v=1}^m w_v=1$, $q$ is the hyperparameter to control the distribution of the weights,
\item Nonlinear combination: It uses a nonlinear function in terms of kernels, namely, multiplication, power, and exponentiation,
\item Data-dependent combination: It assigns specific kernel weights for each data instance, which can identify the local distributions in the data and learn proper kernel combination rules for different regions.
\end{itemize}

\subsubsection{Kernel K-Means and Spectral Clustering}\label{kkmeansc}
Kernel k-means~\cite{bernhar1998nc} and spectral clustering~\cite{inderjit2007pami} are two kernel-based clustering methods for optimizing the intra-cluster variance. Let $\phi(\cdot):\bm x\in\mathcal{\bm X}\rightarrow\mathcal{\bm H}$ be a feature mapping which maps $\bm x$ onto a RKHS $\mathcal{\bm H}$. The kernel k-means method is formulated as the following optimization problem,
\begin{equation}
\begin{split}\label{kkmeans}
\left\{\begin{matrix}
&\underset{ \bm H }{\mbox{min}} \sum_{i=1}^{N}\sum_{k=1}^K H_{ik}\twonorm{\phi(\bm x_i)-\bm \mu_k}^2\\
&s.t.\quad \sum_{k=1}^K H_{ik}=1,\\
\end{matrix}\right.
\end{split}
\end{equation}
where $\bm H\in\{0,1\}^{N\times K}$ is the cluster indicator matrix (also known as cluster assignment matrix), $n_k=\sum_{i=1}^N H_{ik}$ and $\bm \mu_k=\frac{1}{n_k}\sum_{i=1}^N H_{ik}\phi(\bm x_i)$ are the number of points in the $k$th cluster and the centroid of the $k$th cluster. With a kernel matrix $\bm K$ whose $(i,j)$th entry is $K_{ij}={\phi(\bm x_i)}^{\mathrm{T}}\phi(\bm x_j)$, $\bm L=\mbox{diag}([n_1^{-1},n_2^{-1},\cdots,n_K^{-1}])$ and $\bm 1_{l}\in \mathbb{R}^{l}$, a column vector of all ones, Eq.~\eqref{kkmeans} can be equivalently rewritten as the following matrix-vector form,
\begin{equation}
\begin{split}\label{kkmeansmatrixform}
\left\{\begin{matrix}
&\underset{ \bm H }{\mbox{min}}\quad tr(\bm K)-tr({\bm L}^{\frac{1}{2}}\bm H^{\mathrm{T}}\bm K\bm H {\bm L}^{\frac{1}{2}})\\
&s.t.\quad \bm H \bm 1_k=\bm 1_N.\\
\end{matrix}\right.
\end{split}
\end{equation}

For the above kernel k-means matrix-factor form, the matrix $\bm H$ is binary, which makes the optimization problem difficult to solve. By relaxing the matrix $\bm H$ to take arbitrary real values, the above problem can be approximated. Specifically, by defining $\bm U=\bm H{\bm L}^{\frac{1}{2}}$ and letting $\bm U$ take real values, further considering $\mbox{Tr}(\bm K)$ is constant, Eq.~\eqref{kkmeansmatrixform} will be relaxed to
\begin{equation}
\begin{split}\label{sckernelform}
\left\{\begin{matrix}
&\underset{ \bm U }{\mbox{max}}\quad tr(\bm U^{\mathrm{T}}\bm K \bm U)\\
&s.t.\quad \bm U^{\mathrm{T}}\bm U =\bm 1_K.\\
\end{matrix}\right.
\end{split}
\end{equation}
The fact $\bm H^{\mathrm{T}}\bm H={\bm L}^{-1}$ leads to the orthogonality constraint on $\bm U$ which tells us that the optimal $\bm U$ can be obtained by the top $K$ eigenvectors of the kernel matrix $\bm K$. Therefore, Eq.~\eqref{sckernelform} can be considered as the generalized optimization formulation of spectral clustering. Note that Eq.~\eqref{sckernelform} is equivalent to Eq.~\eqref{sc} if the kernel matrix $\bm K$ takes the normalized Gram matrix form.
\subsubsection{Multi-Kernel Based Multi-View Clustering}
Assume that there are $m$ kernel matrices available, each of which corresponds to one view. To make a full use of all views, the weighted combination $\bm K=\sum_{v=1}^m w_v^p \bm K^{(v)}, w_v\geq0,\sum_{v=1}^m w_v=1, p\geq1$ will be used in kernel k-means~\eqref{sckernelform} and spectral clustering~\eqref{sc} to obtain the corresponding multi-view kernel k-means and multi-view spectral clustering~\cite{grigorios2012icdm}. Using the same nonlinear combination but specifically setting $p=1$, Guo et al.~\cite{dongyan2014icpr} extended the spectral clustering to multi-view clustering by further employing the kernel alignment. Due to the potential redundance of the selected kernels, Liu et al.~\cite{xinwang2016aaai} introduced a matrix-induced regularization to reduce the redundancy and enhance the diversity of the selected kernels to attain the final goal of boosting the clustering performance. By replacing the original Euclidean norm metric in fuzzy c-means with a kernel-induced metric in the data space and adopting the weighted kernel combination, Zhang et al.~\cite{daoqiang2003npl} successfully extended the fuzzy c-means to multi-view clustering that is robust to noise and outliers.  In the case when incomplete multi-view data set exists, by optimizing the alignment of the shared data instances, Shao et al.~\cite{weixiang2013icdm} collectively completes the kernel matrices of incomplete data sets. To overcome the cluster initialization problem associated with kernel k-means, Tzortzis et al.~\cite{grigorios2009tnn} proposed a global kernel k-means algorithm, a deterministic and incremental approach that adds one cluster in each stage, through a global search procedure consisting of several executions of kernel k-means from suitable initiations.
\subsubsection{Others}
Besides multi-kernel based multi-view clustering, there are some other methods that use the direct combination of features to perform multi-view clustering like ~\cite{jinlin2016cvpr,feiping2017ijcai}. In~\cite{xiaojun2011tkde}, two-level weights: view wights and variable wights are assigned to the clustering algorithm for multi-view data to identify the importance of the corresponding views and variables. To extend fuzzy clustering method to multi-view clustering, each view is weighted and the multi-view versions of fuzzy c-means and fuzzy k-means are obtained in~\cite{guillaume2009icdm} and~\cite{yizhang2015tc}, respectively.
\subsection{Combination After Projection (Mainly CCA-Based Multi-View Clustering)}
\label{subsec:combination after projection}
For multi-view data  with all views have the same data type like categorical or continuous, it is reasonable to directly combine them together. However, in real-world applications, the multiple representations may have different data types and it is hard to compare them directly. For instance, in bioinformatics, genetic information can be one view while clinical symptoms can be another view in a cluster analysis of patients~\cite{javon2015}. Obviously, the information cannot be combined directly. Moreover, high dimension and noise are difficult to handle. To solve the above problems, the last yet important combination way is introduced: combination after projection. The most commonly used technique is Canonical Correlation Analysis (CCA) and the kernel version of CCA (KCCA).
\subsubsection{CCA and KCCA}
To better understand this style of view combination, CCA and KCCA are briefly introduced (refer to~\cite{david2004nc} for more detail). Given two data sets $\bm S_x=[\bm x_1, \bm x_2,\cdots,\bm x_N]\in {\mathbb{R}}^{d_x\times N}$ and $\bm S_y=[\bm y_1, \bm y_2,\cdots,\bm y_N]\in{\mathbb{R}}^{d_y\times N}$ where each entry $\bm x$ or $\bm y$ has a zero mean, CCA aims to find a projection $\bm w_x\in {\mathbb{R}}^{d_x}$ for $\bm x$ and another projection $\bm w_y\in {\mathbb{R}}^{d_y}$ for $\bm y$ such that the correlation between the projection of $\bm S_x$ and $\bm S_y$ on $\bm w_x$ and $\bm w_y$ are maximized,
\begin{equation}\label{cca1}
\begin{aligned}
\rho=\underset{\bf w_x,\bf w_y}{\mbox{max}}\frac{{\bf w_x}^{\mathrm{T}}\bf C_{xy}\bf w_y}{\sqrt{({\bf w_x}^{\mathrm{T}}\bf C_{xx}\bf w_x)({\bf w_y}^{\mathrm{T}}\bf C_{yy}\bf w_y)}}
\end{aligned}
\end{equation}
where $\rho$ is the correlation and $\bf C_{xy}=\mathbb{E}[\bm x \bm y^{\mathrm{T}}]$ denotes the covariance matrix of $\bm x$ and $\bm y$  with zero mean. Observing that $\rho$ is not affected by scaling $\bf w_x$ or $\bf w_y$ either together or independently, CCA can be reformulated as
\begin{equation}\label{cca2}
\left
\{
\begin{aligned}
&\underset{\bf w_x, \bf w_y}{\mbox{max}}\ \ {\bf w_x}^{\mathrm{T}}\bf C_{xy}\bf w_y\\
& s.t.\ \ \ \ \ {\bf w_x}^{\mathrm{T}}\bf C_{xx}\bf w_x=1,\\
& \ \ \ \ \ \ \ \ \ {\bf w_y}^{\mathrm{T}}\bf C_{yy}\bf w_y=1.
\end{aligned}
\right.
\end{equation}
which can be solved using the method of Lagrange multiplier. The two Lagrange multipliers $\lambda_x$ and $\lambda_y$ are equal to each other, that is $\lambda_x=\lambda_y=\lambda$. If $\bf C_{yy}$ is invertible, $\bf w_y$ can be obtained as $\bf w_y=\frac{1}{2}{\bm C_{yy}}^{-1}\bm C_{yx}\bm w_x$ and $\bf C_{xy}(\bf C_{yy})^{-1}\bf C_{yx}\bf w_x={\lambda}^2\bf C_{xx}\bf w_x$. Hence, $\bf w_x$ can be obtained by solving an eigen problem. For different eigen values (from large to small), eigen vectors are obtained in a successive process.

The above canonical correlation problem can be transformed into a distance minimization problem. For ease of derivation, the successive formulation of the canonical correlation is replaced by the simultaneous formulation of the canonical correlation. Assume that the number of projections is $p$, the matrices $\bf W_x$ and $\bf W_y$ denote $(\bf w_{x1},\bf w_{x2},...,\bf w_{xp})$ and $(\bf w_{y1},\bf w_{y2},...,\bf w_{yp})$, respectively.  The formulation that simultaneously identifies all the $\bf w$'s can be written as an optimization problem with p iteration steps:
\begin{equation}\label{ccasimul}
\left
\{
\begin{aligned}
& \underset{(\bf w_{x1},\bf w_{x2},...,\bf w_{xp}),(\bf w_{y1},\bf w_{y2},...,\bf w_{yp})}{\mbox{max}}\sum_{i=1}^p {\bf w_{xi}}^{\mathrm{T}}\bf C_{xy}\bf w_{yi}\\
& s.t. \ \ \ \ \ \ \ {\bf w_{xi}}^{\mathrm{T}}\bf C_{xx}\bf w_{xj}=\left\{\begin{matrix}
1\ \ \ \ \ \ \ \ \ \mbox{if i=j},\\
0\ \ \mbox{otherwise},
\end{matrix}\right.\\
& \ \ \ \ \ \ \ \ \ \ \ {\bf w_{yi}}^{\mathrm{T}}\bf C_{yy}\bf w_{yj}=\left\{\begin{matrix}
1\ \ \ \ \ \ \ \ \ \mbox{if i=j},\\
0\ \ \mbox{otherwise},
\end{matrix}\right.\\
& \ \ \ \ \ \ \ \ \ \ \ i,j=1,2,...,p,\\
& \ \ \ \ \ \ \ \ \ \ \ {\bf w_{xi}}^{\mathrm{T}}\bf C_{xy}\bf w_{yj}=0,\\
& \ \ \ \ \ \ \ \ \ \ \ i,j=1,2,...,p,j\neq i.\\
\end{aligned}
\right.
\end{equation}

The matrix formulation to the optimization problem~\eqref{ccasimul} is
\begin{equation}\label{ccasimulmatrix}
\left
\{
\begin{aligned}
&\underset{\bf W_x,\bf W_y}{\mbox{max}}\mbox{Tr}({\bf W_x}^{\mathrm{T}}\bf C_{xy}\bf W_y)\\
& s.t.\ \ \ \ \ {\bf W_x}^{\mathrm{T}}\bf C_{xx}\bf W_x=\bf I,\\
&\ \ \ \ \ \ \ \ \ {\bf W_y}^{\mathrm{T}}\bf C_{yy}\bf W_y=\bf I,\\
& \ \ \ \ \ \ \ \ \ {\bf w_{xi}}^{\mathrm{T}}\bf C_{xy}w_{yj}=0,\\
& \ \ \ \ \ \ \ \ \ {\bf w_{yi}}^{\mathrm{T}}\bf C_{yx}w_{xj}=0,\\
&\ \ \ \ \ \ \  \ \ i,j=1,...,p,\ j\neq i.
\end{aligned}
\right.
\end{equation}
where $\bm I$ is an identity matrix with size $p\times p$. Maximizing the objective function of Eq.~\eqref{ccasimulmatrix} can be transformed into the equivalent form as follows:
\begin{equation}\label{ccaequform}
\underset{\bf W_x,\bf W_y}{\mbox{min}}\left \|{\bf W_ x}^{\mathrm{T}} \bf S_x-{\bf W_ y}^{\mathrm{T}}\bf S_y\right \|_F,
\end{equation}
which is used widely in many works~\cite{shaoyuan2014ijcai,qiyue2015cikm,guoqing2016ins}.

KCCA uses the ``kernel trick" to maximize the correlation between two non-linear projected variables. Analogous to Eq.~\eqref{cca2}, the optimization problem for KCCA is formulated as follows:
\begin{equation}\label{kcca}
\left
\{
\begin{aligned}
&\underset{\bf w_x,\bf w_y}{\mbox{max}}\frac{{\bf w_x}^{\mathrm{T}}\bf K_{x}\bf K_y\bf w_y}{\sqrt{({\bf w_x}^{\mathrm{T}}\bf K_{x}^2\bf w_x)({\bf w_y}^{\mathrm{T}}\bf K_{y}^2\bf w_y)}}\\
& s.t.\ \ \ \ \ {\bf w_x}^{\mathrm{T}}\bf K_{x}\bf w_x=1,\\
& \ \ \ \ \ \ \ \ \ {\bf w_y}^{\mathrm{T}}\bf K_{y}\bf w_y=1.
\end{aligned}
\right.
\end{equation}
In contrast to the linear CCA that works by solving an eigen-decomposition of the covariance matrix, KCCA solves the following eigen-problem:
\begin{equation}\label{kccasolution}
\begin{pmatrix}
0 & \bm K_x \bm K_y\\
\bm K_y \bm K_x & 0
\end{pmatrix}\begin{pmatrix}
\bm w_x\\
\bm w_y
\end{pmatrix}=\lambda\begin{pmatrix}
\bm K_x^2 & 0\\
0 & \bm K_y^2
\end{pmatrix}\begin{pmatrix}
\bm w_x\\
\bm w_y
\end{pmatrix}.
\end{equation}
\subsubsection{CCA Based Multi-View Clustering}
Since cluster analysis in a high dimensional space is difficult, Chaudhuri et al.~\cite{kamalika2009icml} firstly projects the data into a lower dimensional space via CCA and then clusters samples in the projected low dimensional space. Under the assumption that multiple views are uncorrelated given the cluster labels, it shows a weaker separation condition required to guarantee the algorithm successful. Blaschko et al.~\cite{matthew2008cvpr} projects the data onto the top directions obtained by the KCCA across different views and applies k-means to clustering the projected samples.

For the case of paired views with some class labels, CCA can still be applied ignoring the class labels, however, the performance can be ineffective. To take an advantage of the class label information, Rasiwasia et al.~\cite{nikhil2014icml} has proposed two solutions with CCA: mean-CCA and cluster-CCA. Consider two data sets each of which is divided into $K$ different but corresponding classes or clusters. Given $\bm S_x=\{\bm x_1, \bm x_2, \cdots, \bm x_K\}$ and  $\bm S_y=\{\bm y_1, \bm y_2,\cdots, \bm y_K\}$, where $\bm x_k=\{\bm x_1^k, \bm x_2^k,\cdots, \bm x_{|\bm x_k|}^k\}$ and $\bm y_k=\{\bm y_1^k, \bm y_2^k,\cdots, \bm y_{|\bm y_k|}^k\}$ are the data points in the $k$th cluster for the first and second views, respectively. The first solution is to establish correspondences between the mean cluster vectors in the two views. Given the cluster means $\bm m_x^k=\frac{1}{|\bm x_k|}\sum_{i=1}^{|\bm x_k|}x_i^k$ and $\bm m_y^k=\frac{1}{|\bm y_k|}\sum_{i=1}^{|\bm y_k|}y_i^k$, mean-CCA is formulated as
\begin{equation}
 \rho=\underset{\bm w_x,\bm w_y}{\mbox{max}}\frac{\bm w_x\bm V_{xy}\bm w_y}{\sqrt{({\bf w_x}^{\mathrm{T}}\bf V_{xx}\bf w_x)({\bf w_y}^{\mathrm{T}}\bf V_{yy}\bf w_y)}},
\end{equation}
where $\bm V_{xy}=\frac{1}{K}\sum_{k=1}^K\bm m_x^k{\bm m_y^k}^{\mathrm{T}}$, $\bm V_{xx}=\frac{1}{K}\sum_{k=1}^K\bm m_x^k{\bm m_x^k}^{\mathrm{T}}$ and $\bm V_{yy}=\frac{1}{K}\sum_{k=1}^K\bm m_y^k{\bm m_y^k}^{\mathrm{T}}$. The second solution is to establish a one-to-one correspondence between all pairs of data points in a given cluster across the two views of data sets and then standard CCA is used to learn the projections.

For multi-view data with at least one complete view (features for this view are available for all data points), Anusua et al.~\cite{trivedi2010} borrowed the idea from Laplacian regularization to complete the incomplete kernel matrix and then applied KCCA to perform multi-view clustering. In another method for multi-view clustering, multiple pattern matrices $\bm A^{(v)}\in {\mathbb{R}}^{N\times K_v}, v=1, 2,\cdots, K$ each of which corresponds to a view are obtained in an intermediate step and then a consensus pattern matrix should be learned to approximate each view's pattern matrix as much as possible. Due to the unsupervised property, however, the pattern matrices are often not directly comparable. Using the CCA formulation Eq.~\eqref{ccaequform}, Long et al.~\cite{bo2008sdm}  projects one view's pattern matrix first before comparing with another view's pattern matrix. 

The same idea can be used to tackle the incomplete view problem (i.e., there are no complete views). For instance, if there are only two views, the methods in~\cite{shaoyuan2014ijcai,qiyue2015cikm} split data into the portion of data with both views and the portion of data with only one view, and then projects each view's data matrix so that it is close to the final indicator matrix. Multi-view information is connected by the common indicator matrix corresponding to the projected data from both views. Wang et al.~\cite{qiang2016neurocomputing} provides a multi-view clustering method using an extreme learning machine that maps the normalized feature space onto a higher dimensional feature space.

\subsection{Discussion}
\label{subsec:discussion}

\begin{figure*}[!htbp]
\begin{center}
\centerline{\includegraphics[width=2\columnwidth,height=0.8\columnwidth]{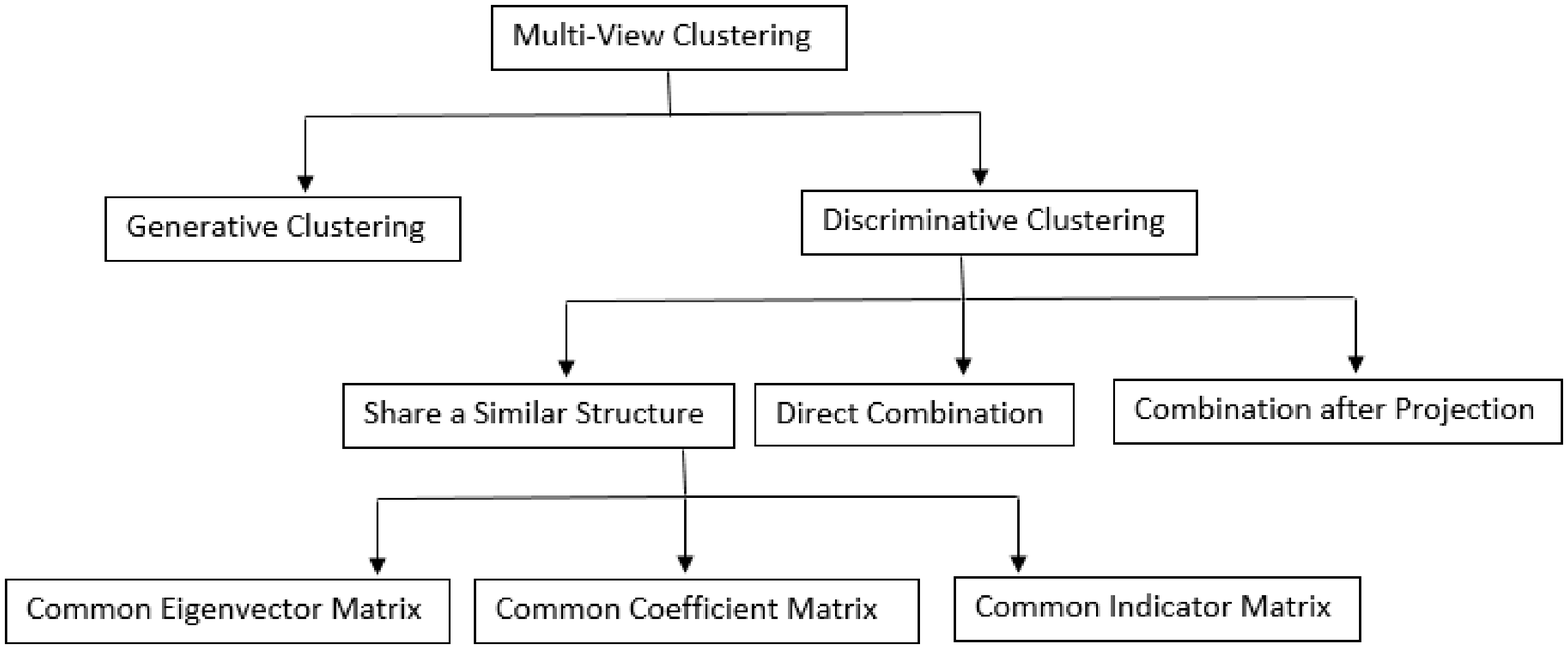}}
\centering
\caption{The taxonomy of multi-view clustering methods.}\label{fig:structure}
\end{center}
\vskip -0.4in
\end{figure*}

In Fig.~\ref{fig:structure}, we give the taxonomy of the multi-view clustering methods, which is also how this survey is organized. Now, we give some discussions about these methods displayed in Fig~\ref{fig:structure}. For multi-view generative clustering, there are two advantages: first,  it can deal with missing values naturally; second, some convex model can obtain the global solution.  However, there are two disadvantages acompanied: first, it is based on some user assumption that may be false, thus resulting in inaccurate cluster results ;  second, it is time consuming because it introduces some model parameters and needs to run different executions for convex model. For multi-view discriminative clustering, three  classes of similarity structure shared methods make good use of the multi-view consensus information, but in some situation the similarity structure may be too strict.  Common eigenvector matrix shared method is based on the spectral clustering, which applies to any shape clusters. Common coefficient matrix shared method  mainly includes the subspace clustering, which is extensively used in computer vision field. Common indicator matrix shared method mainly includes k-means and non-negative matrix factorization, thus it has vast variety of applications. Direct combination based method can adaptively tune the weights of each view, which is in need when some views are low-quality. Combination after projection works in the scenario where different views cannot be directly compared in original space.  It is difficult to claim which one is better, it depends on the specific application.

In Fig.~\ref{fig:structure},  we can find that for the three classes of multi-view clustering methods introduced in subsections ~\ref{subsec:multi-view spectral clustering},~\ref{subsec:multi-view subspace clustering},~\ref{subsec:multi-view nonnegative matrix factorization clustering}, in fact, a common property is that these methods combine multiple views by sharing a similar structure across the  multiple views. There are also some methods to share other similar structures to perform multi-view clustering. By sharing an indicator vector across views in a singular value decomposition of multiple data matrices, Sun et al.~\cite{javon2015,javon2013,javon2014} extend the bi-clustering~\cite{mihee2010biometrics} method to the multi-view settings. Wang et al.~\cite{chang2016tkde} chooses the Jaccard similarity to measure the cross-view clustering consistency and simultaneously considers the within-view clustering quality to cluster multi-view data.

Besides these categorized methods, there are some other multi-view clustering methods. Different from exploiting the consensus information of multi-view data, Cao et al.~\cite{xiaochun2015} utilizes a Hilbert Schmidt Independence Criterion as a diversity term to explore the complementarity of multi-view information. It reduces the redundancy of multi-view information to improve the clustering performance. Based on the idea of ``minimizing disagreement" between clusters from each view, De Sa~\cite{virginia2005icmlworkshop} proposes a two-view spectral clustering that creates a bipartite graph of the views. Zhou et al.~\cite{dengyongicml2007} defines a mixture of Markov chains on similarity graph of each view and generalize spectral clustering to multiple views. In~\cite{rongkai2014aaai}, a transition probability matrix is constructed from each single view, and all these transition probability matrices are used to recover a shared low-rank transition probability matrix as a crucial input to the standard Markov chain method for clustering. By fusing the similarity data from different views, Lange et al.~\cite{tilman2005nips} formulates a nonnegative matrix factorization problem and adopts an entropy-based mechanism to control the weights of multi-view data. Liu et al.~\cite{xinhai2012tkde} chooses tensor to represent multi-view data and then performs cluster analysis via tensor methods.

\section{Relationships to Related Topics}
\label{sec:relationships to related topics}
As we mentioned previously, MVC is a learning paradigm for cluster analysis with multi-view feature information. It is a basic task in machine learning and thus can be useful for various subsequent analyses. In machine learning and data mining fields, there are several closely related learning topics such as multi-view representation learning, ensemble clustering, multi-task clustering, multi-view supervised and semi-supervised learning. In the following, we will elaborate the relationships between MVC and a few other topics.
%\subsection{Relationship to Multi-View Representation}
%\label{subsec:relationship to multi-view representation}

Multi-view representation~\cite{yingming2016} is the problem of learning a more comprehensive or meaningful representation from multi-view data. According to~\cite{yoshua2012arxiv}, representation learning (also named feature engineering) is a way to take advantage of human ingenuity and prior knowledge to extract some useful but far-removed feature representation for the ultimate objective. Representation learning is also unsupervised, which is the same as clustering in the sense that they do not use label information.  Multi-view representation can be considered as a more basic task than multi-view clustering, since multi-view representation can be useful in broader purpose such as classification or clustering and so on. However, cluster analysis based on multi-view representation may not be ideal because the creation of multi-view representation is unaware of the final goal of clustering. In an archived survey article~\cite{yingming2016}, multi-view representation methods are categorized into mainly two classes: the shallow methods and the deep methods. The shallow methods are mainly based on CCA, which may correspond to our subsection~\ref{subsec:combination after projection}. For the deep methods, there exist a large number of works~\cite{nitish2015jmlr,jiquan2011icml,junhua2014,fangxiang2015mm,weiran2015icml,andrej2017pami,jeff2017pami} on multi-view representation. However, for multi-view deep clustering, there are only a few including~\cite{fei2014aaai,junyuan2016icml}. As mentioned above, the sequential way of first multi-view representation and then clustering is a natural way to perform multi-view clustering, but the ultimate performance is usually not good because of the gap in the two steps. Therefore, how to integrate clustering and multi-view representation learning into a simultaneous process is an intriguing direction up to date, especially for deep multi-view representation.
%\subsection{Relationship to Ensemble Clustering}
%\label{subsec:relationship to ensemble clustering}

Ensemble clustering~\cite{sandro2011ijprml} (also named consensus clustering or aggregation of clustering) is  to reconcile clustering information about the same data set coming from different sources or from different runs of the same clustering method to find a single consensus clustering that is a better fit in some sense than any one else in the ensemble. If ensemble clustering is applied to clustering with multiple views of data, it becomes a type of multi-view clustering method. Therefore, all of the ensemble clustering techniques e.g.,~\cite{xiaoli2004icml,mete2013icip,eric2013bioinformatics,yasin2014sciencereports,hongfu2017tkde} can be applied to MVC. For instance, ~\cite{zhiqiang2017ijcai,xijiong2013icmlc} are two multi-view ensemble clustering methods.
\begin{figure*}[!htbp]\label{}
\begin{center}
\centerline{\includegraphics[width=2\columnwidth,height=0.8\columnwidth]{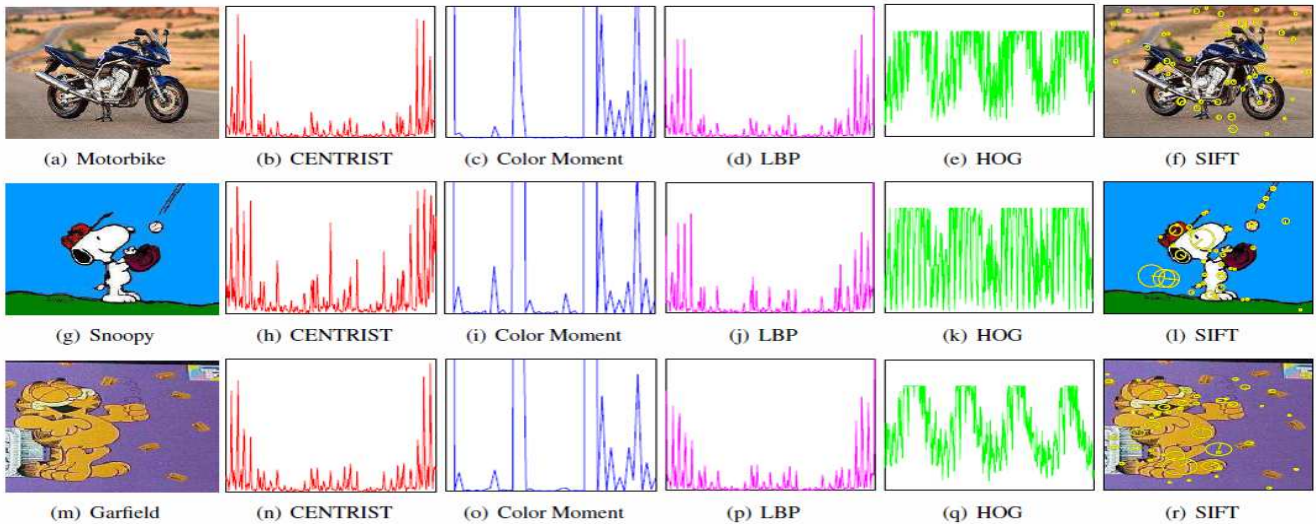}}
\centering
\caption{The five views (CENTRIST, ColorMoment, LBP, HOG and SIFT) on three sample images from Caltech101.}
\label{fig:cv}
\end{center}
\vskip -0.4in
\end{figure*}

Multi-task clustering aims to improve the performance of unsupervised clustering tasks, such as~\cite{jianwen2011nc,xianchao2015tkdd,quanquan2011aaai,xiaolei2015pami,xianchao2016ijcai}. If each task corresponds to clustering in a specific view of the same sample, multiple clustering results will be obtained, and then ensemble clustering methods may be employed to fuse these clustering results. Therefore, multi-task clustering potentially combined with ensemble clustering can implement multi-view clustering. In addition, multi-task clustering and multi-view clustering can be conducted simultaneously to improve the clustering performance~\cite{xianchao2015ijcai,xiaochao2016tkde}.
%\subsection{Relationship to Multi-View Supervised and Semi-Supervised Learning}
%\label{subsec:relationship to multi-view supervised and semi-supervised learning}

Different from multi-view clustering, multi-view supervised learning~\cite{sunshilang2013} uses the labeled data to learn classifiers (or other inference models) while multi-view semi-supervised learning~\cite{xu2013,sunshilang2013} can learn classifiers with both the labeled and unlabeled data. The commonality between them lies in the way to combine multiple views. Many widely recognized techniques for combining views in the supervised or semi-supervised settings, e.g., co-training~\cite{blum1998combining,yu2011bayesian}, co-regularization~\cite{Sindhwani05aco-regularized,vikas2008icml}, margin consistency~\cite{sls2013ijcai,guoqing2016tnnls} can lend a hand to multi-view clustering if there is a mechanism to estimate the initial labels.
\section{Applications}
\label{sec:applications}

Multi-view clustering has been successfully applied to various applications including computer vision, natural language processing, social multimedia, bioinformatics and health informatics and so on.
\subsection{Computer Vision}
\label{subsec:cv}
Multi-view clustering has been widely used in image categorization~\cite{qiyue2015,yang2015,hongchang2015,xiaochun2015,cheng2015aaai,mete2013icip,andres2015ist} and motion segmentation tasks~\cite{zeynep2011,abdelaziz2013iccv}. Typically, several feature types e.g., CENTRIST~\cite{jianxin2011pami}, ColorMoment~\cite{hui2002icip}, HOG~\cite{navneet2005cvpr}, LBP~\cite{timo2002pami} and SIFT~\cite{david2004ijcv} can be extracted from the images (see the Fig.~\ref{fig:cv}~\cite{hongchang2015}) prior to cluster analysis. Yin et al.~\cite{qiyue2015} proposed a pairwise sparse subspace representation for multi-view image clustering, which harnesses the prior information and maximizes the correlation between the representations of different views. Wang et al.~\cite{yang2015} enforced between-view agreement in an iterative way to perform multi-view spectral clustering on images. Gao et al.~\cite{hongchang2015} assumed a common low dimensional subspace representation for different views to reach the goal of multi-view clustering in computer vision applications. Cao et al.~\cite{xiaochun2015} adopted Hilbert Schmidt Independence Criterion as a diversity term to exploit the complementary information of different views and performed well on both image and video face clustering tasks. Jin et al.~\cite{cheng2015aaai} utilized the CCA to perform multi-view image clustering for large-scale annotated image collections.

Ozay et al. ~\cite{mete2013icip} used consensus clustering to fuse image segmentations. M\'{e}ndez et al.~\cite{andres2015ist} adopted the ensemble way to perform multi-view clustering for MRI image segmentation.  Nonnegative matrix factorization was adopted in ~\cite{zeynep2011} to perform multi-view clustering for motion segmentation. Djelouah et al.~\cite{abdelaziz2013iccv} addressed the motion segmentation problem by propagating segmentation coherence information in both space and time.
\subsection{Natural Language Processing}
\label{subsec:nlp}
In natural language processing, text documents can be obtained in multiple languages. It is natural to use multi-view clustering to conduct document categorization~\cite{Kumar2011ICML,Kumar2011NIPS,jialu2013,hongchang2015,youngmin2010sigir,yu2012icpr} with each language as one view. Employing the co-training and co-regularization ideas, Kumar et al.~\cite{Kumar2011ICML,Kumar2011NIPS} proposed co-training multi-view clustering and co-regularization multi-view clustering, respectively. The performance comparison on multilingual data demonstrates the superiority of these two methods over single-view clustering. Liu et al.~\cite{jialu2013} extended nonnegative matrix factorization to multi-view settings for clustering multilingual documents. Kim et al.~\cite{youngmin2010sigir} obtained the clustering results from each view and then constructed a consistent data grouping by voting. Jiang et al.~\cite{yu2012icpr} proposed a collaborative PLSA method that combines individual PLSA models in different views and imports a regularizer to force the clustering results in an agreement across different views. Hussain~\cite{syed2014iis} utilized an ensemble way to perform multi-view clustering on documents.

\subsection{Social Multimedia}
\label{subsec:social multimedia}
\begin{figure}[!htbp]
\begin{center}
\centerline{\includegraphics[width=\columnwidth, height=0.4\columnwidth]{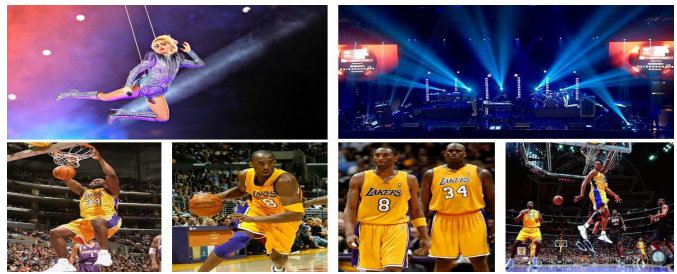}}
\centering
\caption{Some pictures from two social events: concerts (top row) and NBA game (bottom row).}
\label{fig:sm}
\end{center}
\vskip -0.4in
\end{figure}

Currently, with the fast development of social multimedia, how to make full use of large quantities of social multimedia data is a challenging problem, especially match them to the ``real-world concepts" such as the ``social event detection". Fig.~\ref{fig:sm} shows two such events: a concert, and an NBA game. The pictures showed there form just one view, and other textural features such as tags and titles form the other view. Such a social event detection problem is a typical multi-view clustering problem. Petkos et al.~\cite{georgios2012icmr} adopted a multi-view spectral clustering method to detect the social event and additionally utilized some known supervisory signals (the known clustering labels). Samangooei et al.~\cite{Samangooei2013SocialED} performed feature selection first before constructing the similarity matrix and applied a density based clustering to the fused similarity matrix. Petkos et al.~\cite{georgios2014icmm} proposed a graph-based multi-view clustering to cluster the data from social multimedia. Multi-view clustering has also been applied to grouping multimedia collections~\cite{ron2007cvpr} and news stories~\cite{xiao2008tm}.

\subsection{Bioinformatics and Health Informatics}
\label{subsec:BIHI}
\begin{figure}[!htbp]
\begin{center}
\centerline{\includegraphics[width=\columnwidth, height=0.3\columnwidth]{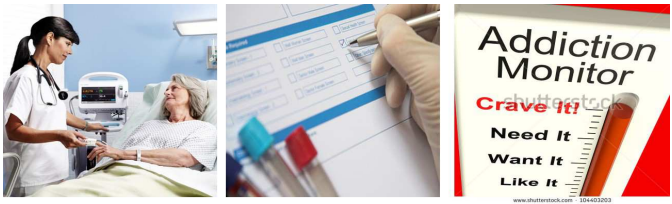}}
\centering
\caption{Three views from health informatics: vital sign (left), urine drug screen (middle) and craving measure (right)).}
\label{fig:hi}
\end{center}
\vskip -0.4in
\end{figure}

In order to identify genetic variants underlying the risk for substance dependence, Sun et al.~\cite{javon2015,javon2013,javon2014} designed three multi-view co-clustering methods to refine diagnostic classification to better inform genetic association analyses. Chao et al.~\cite{guoqing2018tcbb} extended the method in \cite{javon2015} to handle missing values that might appear in every view of the data, and used the method to analyze heroin treatment outcomes. The three views of data for heroin dependence patients are demonstrated in Fig.~\ref{fig:hi}. Yu et al.~\cite{shi2012pami,shi2011bioinformatics} designed a multi-kernel combination to fuse different views of information and showed superior performance on disease data sets. In~\cite{dan2016bhi}, a multi-view clustering based on the Grassmann manifold was proposed to deal with gene detection for complex diseases.
\section{Open Problems}
\label{sec:open problems}
We have identified several problems that are still underexplored in the current body of MVC literature. We discuss these problems in this section.
\subsection{Large Scale Problem (size and dimension)}
\label{subsec:large scale}
In modern life, large quantities of data are generated every day. For instance, several million posts are shared per minute in Facebook, which include multiple data forms (views): videos, images and texts. At the same time, a large amount of news are reported in different languages, which can also be considered as multi-view data with each language as one view.  However, most of the existing multi-view clustering methods can only deal with small datasets. It is important to extend these methods to large scale applications. For instance, it is difficult for the existing multi-view spectral clustering based methods to work on datasets of massive samples due to the expensive computation of graph construction and eigen-decomposition. Although some previous works such as~\cite{deng2014tc,charless2004pami,donghui2006ecml,donghui2009kdd} attempted to accelerate the spectral clustering method to scale with big data, it is intriguing to extend them effectively to the multi-view settings.

Another type of big data has high dimensionality. For instance, in bioinformatics, each person has millions of genetic variants as genetic features where compared with the problem dimension, the number of samples is low. Using genetic features in a clinical analysis  with another view of clinical phenotypes, it often forms multi-view analytics problem. How to deal with such a clustering problems is tough due to the over-fitting problem.  Although feature selection~\cite{jennifer2004jmlr,daniela2010jasa} or feature dimension reduction like PCA is commonly used to alleviate this problem in single-view settings, there are not convincing methods up to now, especially deep learning cannot cope with it due to the  properties: small size and high feature dimension. It may recall new theory to appear to handle this problem.

\subsection{Incomplete Views or Missing Value}
\label{subsec:incomplete views or missing value}

Multi-view clustering has been successfully applied to many applications as shown in Section~\ref{sec:applications}. However, there is an underlying problem hidden behind: what if one or more views are incomplete. This is very common in real-world applications. For example, in multi-lingual documents, many documents may have only one or two language versions; in social multimedia, some sample may miss visual or audio information due to sensor failure; in health informatics, some patients may not take certain lab tests to cause missing views or missing values. Some data entries may be missing at random while others are non-random. Simply replacing the missing entries with zero or mean values~\cite{yoshikazu2002pakdd} is a common way to deal with the missing value problem, and multiple imputation~\cite{su2011} is also a popular method in statistical field. The missing entries can be generated by the recently popular generative adversarial networks~\cite{chao2017bigdata}. However, without considering the differences of random and non-random effects in missing data, the clustering performance is not ideal~\cite{guoqing2018tcbb}.

Up to now, there have already been several multi-view works ~\cite{nishant2016icpr,shaoyuan2014ijcai,handong2016ijcai,weixiang2013icdm,qiyue2015cikm,weixiang2015,weixiang2016bigdata,trivedi2010} that attempted to solve the incomplete view problem. Two methods in~\cite{weixiang2015,weixiang2016bigdata} introduced a weight matrix $M_{i,j}$ to indicate whether the $i$th instance present in the $j$th view. For the two-view case, the method in~\cite{shaoyuan2014ijcai} reorganized the multi-view data to include three parts:  samples with both two views, samples only having view 1 and samples only having view 2 and then analyzed them to handle missing entries. Assuming that there is at least one complete view, Trivedi et al.~\cite{trivedi2010} used the graph Laplacian to complete the kernel matrix with missing values based on the kernel matrix computed from the complete view. Shao~\cite{weixiang2013icdm} borrowed the same idea to deal with multi-view setting. It is noted that all these methods deal with incomplete views or missing value with some constraints, they do not aim to deal with the situation with arbitrarily missing values in any of the views. In other words, this situation is that all views have missing values and the samples just miss a few features in a view. Obviously, the above methods have significant limitations that cannot make full use of the available multi-view incomplete information   In addition, all existing methods do not take into consideration the difference between random and non-random missing patterns. Therefore, it is worth exploring how to use the mixed types of data in multi-view analysis.
\subsection{Local Minima}
\label{subsec:theory}
For multi-view clustering methods based on k-means, the initial clusters are very important and different initalizations may lead to different clustering results. It is still challengig to select the initial clusters effectively in MVC and even in single-view clustering settings. 

Most NMF-based methods rely on non-convex optimization formulations, and thus are prone to the local optimum problem, especially when missing values and outliers exist. Self-paced learning~\cite{qian2015aaai} is a possible solution, and Xu et al.~\cite{chang2015ijcai} applied it to multi-view clustering to alleviate the local minimum problem.

The generative convex clustering method~\cite{danial2008nips} is an interesting approach to avoid the local minimum problem. In~\cite{grigorios2009icann}, a multi-view version of the method in~\cite{danial2008nips} is proposed and shows good performance. This kind of generative methods may be another good solution.

\subsection{Deep Learning}
\label{subsec:deep learning}
Recently, Deep learning has demonstrated outstanding performance in many applications such as speech recognition, image segmentation, object detection and so on. However, there are few deep learning works on clustering, let alone multi-view clustering. The common way in the deep learning paradigm is to learn a good multi-view data representation using a deep model and then apply a regular clustering method to cluster samples based on the resultant data representation.

The works in~\cite{marc2017icml,john2016icassp,hyun2017cvpr} borrowed the supervised deep learning idea to perform supervised clustering. In fact, they can be considered as performing semi-supervised learning. So far, there are only several truly deep clustering works~\cite{fei2014aaai,junyuan2016icml}. Tian et al.~\cite{fei2014aaai} proposed a deep clustering algorithm that is based on spectral clustering, but replaced eigenvalue decomposition by a deep auto-encoder. Xie et al.~\cite{junyuan2016icml} proposed a clustering approach using deep neural network which can learn representation and perform clustering simultaneously. Now, extending these single-view deep clustering methods to multi-view settings or designing multi-view deep clustering methods are promising future directions.
\subsection{Mixed Data Types}
Multi-view data may not necessarily just contain numerical or categorical features. They can also have other types such as symbolic, and ordinal, etc. These different types can appear simultaneously in the same view, or in different views. How to integrate different types of data to perform multi-view clustering is worthy of careful investigation. Converting all of them to categorical type is a straightforward solution. However, much information will be lost during such a processing. For example, the difference of the continuous values categorized into the same category is ignored. It is worth exploring to make full use of the information within mixed data types in multi-view clustering setting.
\subsection{Multiple Solutions}
Most of the existing multi-view clustering, even single-view clustering algorithms only output a single clustering solution. However, in real-world applications, data can often be grouped in many different ways and all these solutions are reasonable and interesting from different perspectives. For example, it is both reasonable to group the fruits apple, banana, and grape according to the fruit type or color.  Until now, to the best of our knowledge, there are only two works along this direction~\cite{ying2007icdm,donglin2010icml}. Cui et al.~\cite{ying2007icdm} proposed to partition multi-view data by projecting the data to a space that is orthogonal to the current solution so that multiple non-redundant solutions were obtained. In another work~\cite{donglin2010icml}, Hilbert-Schmidt Independence Criterion was adopted to measure the dependence across different views and then one clustering solution was found in each view.  Multi-view clustering algorithms that can produce multiple solutions should attract more attentions in the future.
\section{Conclusion}
\label{sec:conclusion}
In this paper, we have reviewed two major types of multi-view clustering methods:  generative methods and discriminative methods. Because of the large variety of discriminative methods, based on the ways that they integrate views, we split them into five main classes, the first three of which have a commonality: sharing certain structures across the views, the fourth one contains direct combinations of the views while the fifth one includes view combinations after projections. As for generative methods, we can find that they have developed far less sufficiently than discriminative ones. To better understand multi-view clustering, we elaborate the relationships between MVC and several closely related learning methods. We have also introduced several real-world applications of MVC and pointed out some interesting and challenging future directions.

% or


\ifCLASSOPTIONcompsoc
  \section*{Acknowledgments}This work was supported by National Institutes of Health (NIH) grants R01DA037349 and K02DA043063, and National Science Foundation (NSF) grants DBI-1356655, CCF-1514357, and IIS-1718738. Jinbo Bi was also supported by NSF grants  IIS-1320586, IIS-1407205, and IIS-1447711.
\else
  \section*{Acknowledgment}This work was supported by National Institutes of Health (NIH) grants R01DA037349 and K02DA043063, and National Science Foundation (NSF) grants DBI-1356655, CCF-1514357, and IIS-1718738. Jinbo Bi was also supported by NSF grants  IIS-1320586, IIS-1407205, and IIS-1447711.
\fi

\ifCLASSOPTIONcaptionsoff
  \newpage
\fi



%\bibliographystyle{IEEEtran}
{\small
\bibliographystyle{IEEEtran}
\bibliography{multiviewclustering}
}


%\end{IEEEbiography}
%% insert where needed to balance the two columns on the last page with
%%\newpage


%\vfill


\end{document}